\documentclass{article}
\usepackage[utf8]{inputenc}
\usepackage{setspace}
\usepackage[margin=1.1in]{geometry}
\usepackage{graphicx}
\graphicspath{ {./figures/} }

\usepackage[utf8]{inputenc} 
\usepackage[T1]{fontenc}    
\usepackage{hyperref}       
\usepackage{url}            
\usepackage{booktabs}       
\usepackage{amsfonts}       
\usepackage{nicefrac}       
\usepackage{microtype}      
\usepackage{xcolor}         
\usepackage{bm}
\usepackage{graphicx}
\usepackage{parskip}
\usepackage{algorithmic}

\usepackage{amsthm}
\newtheorem*{lemma}{Lemma}

\newtheorem{sublemma}{Lemma}
\usepackage{algorithm}
\usepackage{mathtools}
\usepackage{amssymb}
\usepackage{wrapfig}
\usepackage{setspace}
\usepackage{xcolor} 

\usepackage{placeins}
\usepackage{tikz}
\usetikzlibrary{fit,calc}

\usepackage{bbm}

\usepackage{amsmath,amsfonts,bm}

\def\eqref#1{equation~\ref{#1}}

\def\1{\bm{1}}

\def\rmP{{\mathbf{P}}}

\DeclareMathAlphabet{\mathsfit}{\encodingdefault}{\sfdefault}{m}{sl}
\SetMathAlphabet{\mathsfit}{bold}{\encodingdefault}{\sfdefault}{bx}{n}

\newcommand{\E}{\mathbb{E}}

\title{Federating for Learning Group Fair Models}

\author{%
 Afroditi Papadaki \textsuperscript{1*} 
   \and
   Natalia Martinez \textsuperscript{2}
   \and
   Martin Bertran \textsuperscript{2}
   \and
   Guillermo Sapiro \textsuperscript{2} \\
   \and
   Miguel Rodrigues \textsuperscript{1}\\
}

\date{\vspace{-6ex}}

\begin{document}

\maketitle

{\begin{center}
    \textsuperscript{1}University College London, London, UK;\\ \texttt{\{a.papadaki.17, m.rodrigues\}@ucl.ac.uk}\\
\textsuperscript{2} Duke University, Durham, NC, USA;\\
\texttt{\{natalia.martinez, martin.bertran, guillermo.sapiro\}@duke.edu}\\
\textsuperscript{*} Corresponding author
\end{center}}

\begin{abstract}
 
Federated learning is an increasingly popular paradigm that enables a large number of entities to collaboratively learn better models. In this work, we study minmax group fairness in paradigms where different participating entities may only have access to a subset of the population groups during the training phase. We formally analyze how this fairness objective differs from existing federated learning fairness criteria that impose similar performance across participants instead of demographic groups. We provide an optimization algorithm -- FedMinMax -- for solving the proposed problem that provably enjoys the performance guarantees of centralized learning algorithms. We experimentally compare the proposed approach against other methods in terms of group fairness in various federated learning setups.

\end{abstract}


\section{Introduction}

Machine learning models are being increasingly adopted to make decisions in a range of domains, such as finance, insurance, medical diagnosis, recruitment, and many more \cite{10.1145/3376898}. Therefore, we are often confronted with the need -- sometimes imposed by regulatory bodies -- to ensure that such machine learning models do not lead to decisions that discriminate individuals from a certain demographic group.

\begin{wrapfigure}{r}{0.55\textwidth} 
    \centering
    \includegraphics[width=7.8cm]{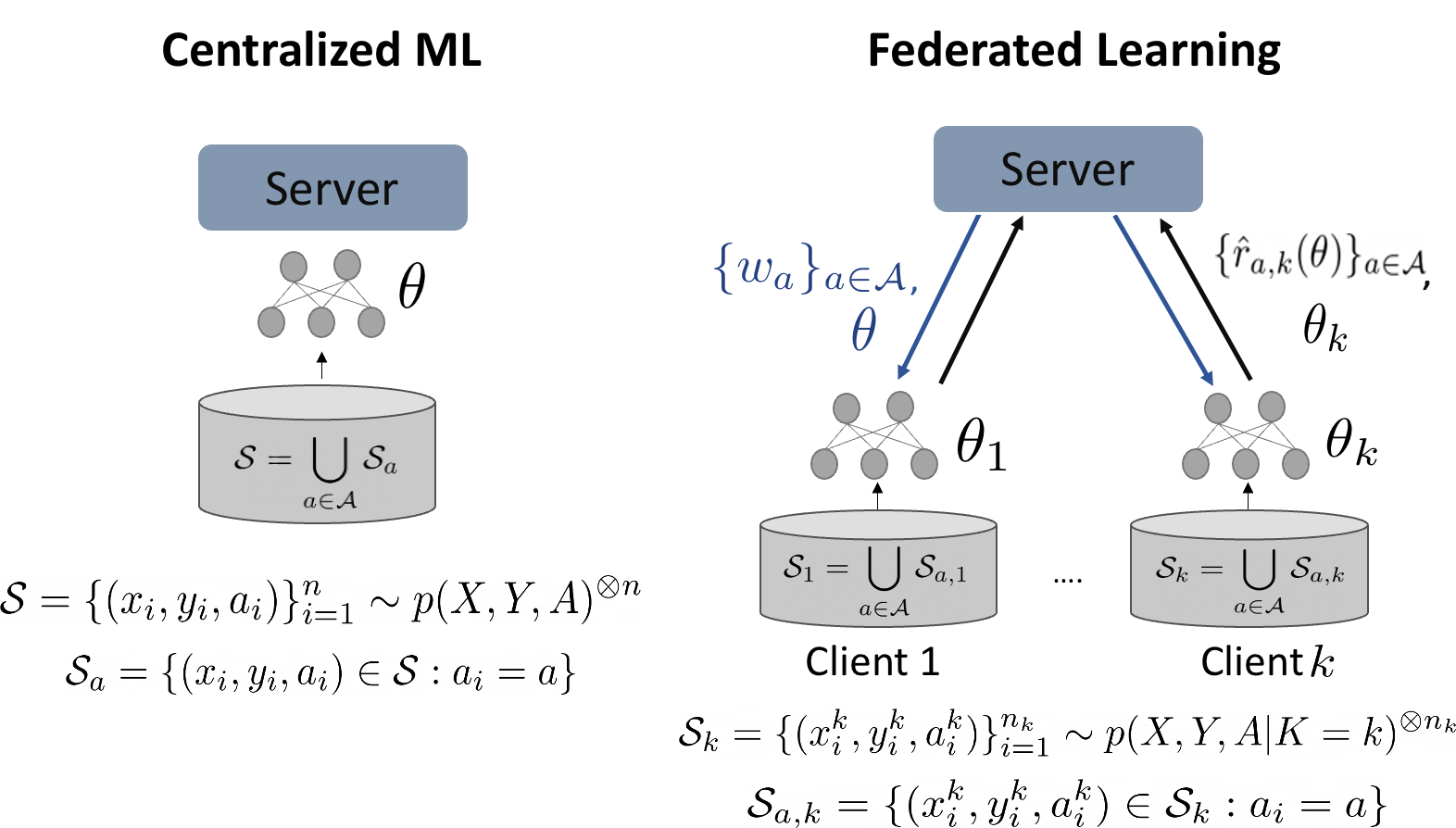}
    
    \caption{Centralized Learning vs. Federated Learning group fairness. \textit{Left:} A single entity holds the dataset $\mathcal{S}$ in a single server that is responsible for learning a model $h$ parameterized by $\theta$.  \textit{Right:} Multiple entities hold different datasets $\mathcal{S}_k$, sharing restricted information with a server that is responsible for learning a model $h$ parameterized by $\theta$. See also Section 3.
    }
    \label{fig:Fairschemes}
\end{wrapfigure}

The development of machine learning models that are fair across different (demographic) groups has been well studied in traditional learning setups where there is a single entity responsible for learning a model based on a local dataset holding data from individuals of the various groups. 
However, there are various settings where the data representing different demographic groups is spread across multiple entities rather than concentrated on a single entity/server. For example, consider a scenario where various hospitals wish to learn a diagnostic machine learning model that is fair (or performs reasonably well) across different demographic groups but each hospital may only contain training data from certain groups because -- in view of its geo-location -- it serves predominantly individuals of a given demographic \cite{DBLP:journals/corr/abs-2108-08435}. 
This new setup along with the conventional centralized one are depicted in Figure \ref{fig:Fairschemes}.

These emerging scenarios however bring about various challenges. The first challenge relates to the fact that each individual entity may not be able to learn locally by itself a fair machine learning model because it may not hold (or hold little) data from certain demographic groups; The second relates to that fact that each individual entity may also not be able to directly share their own data with other entities due to legal or regulatory challenges such as GDPR \cite{EUdataregulations2018}. Therefore, the conventional machine learning fairness \textit{ansatz} -- relying on the fact that the learner has access to the overall data --  does not generalize from the centralized data setup to the new distributed one.

It is possible to partially address these challenges by adopting federated learning (FL) approaches. These learning approaches enable multiple entities (or clients\footnote{Clients are different user devices, organisations or even geo-distributed datacenters of a single company \cite{advancesFL}. In this manuscript we use the terms participants, clients and entities, interchangeably.}) coordinated by a central server to iteratively learn in a decentralized manner a single global model to carry out some task~\cite{DBLP:journals/corr/KonecnyMRR16,DBLP:journals/corr/KonecnyMYRSB16}. The clients do not share data with one another or with the server; instead the clients only share focused updates with the server, the server then updates a global model, and distributes the updated model to the clients, with the process carried out over multiple rounds or iterations. This learning approach enables different clients with limited local training data to learn better machine learning models.

However, with the exception of \cite{DBLP:journals/corr/abs-2108-08435}, which we will discuss later, federated learning is not typically used to learn models that exhibit performance guarantees for different demographic groups served by a client (i.e. \textit{group fairness} guarantees); instead, it is primarily used to learn models that exhibit specific performance guarantees for each client involved in the federation (i.e. \textit{client fairness} guarantees). Importantly, in view of the fact that a machine learning model that is \textit{client fair} is not necessarily \textit{group fair} (as we later demonstrate formally in this work), it becomes crucial to understand how to develop new federated learning techniques leading up to models that are also fair across different demographic groups.

This work develops a new federated learning algorithm that can be adopted by multiple entities coordinated by a single server to learn a global (minimax) group fair model. We show that our algorithm leads to the same (minimax) group fairness performance guarantees of centralized approaches such as \cite{diana2020convergent,martinez2020minimax}, which are exclusively applicable to settings where the data is concentrated in a single client. Interestingly, this also applies to scenarios where certain clients do not hold any data from some of the groups.

The rest of the paper is organized as follows: Section \ref{sec:related_work} overviews related work. Section \ref{sec:prob_form} formulates our proposed distributed group fairness problem. Section \ref{clientvsgroupfairness} formally demonstrates that traditional federated learning approaches such as \cite{DBLP:journals/corr/abs-2108-08435,DRFA,Li2020Fair,AFL} may not always solve group fairness. In Section \ref{sec:algo} we propose a new federated learning algorithm 
to collaboratively learn models that are minimax group fair. Section \ref{sec:experiemtns} illustrates the performance of our approach in relation to other baselines.  
Finally, Section \ref{sec:conclusion} draws various conclusions.

\section{Related Work}\label{sec:related_work}
The development of fair machine learning models in the standard \textit{centralized learning setting} -- where the learner has access to all the data -- is underpinned by fairness criteria. One popular criterion is \textit{individual fairness} \cite{dwork2011fairness} that dictates that the model is fair provided that people with similar characteristics/attributes are subject to similar model predictions/decisions.

Another 
family of criteria -- known as \textit{group fairness} -- requires the model to perform similarly on different demographic groups. Popular  group fairness criteria include equality of odds, equality of opportunity \cite{hardt2016equality}, and demographic parity \cite{louizos2017variational}, that are usually imposed as a constraint within the learning problem.
More recently, \cite{martinez2020minimax} introduced \textit{minimax group fairness}; this criterion requires the model to optimize the prediction performance of the worst demographic group without unnecessarily impairing the performance of other demographic groups (also known as no-harm fairness) \cite{diana2020convergent,martinez2020minimax}. 

The development of fair machine learning models in \textit{federated learning settings} has been building upon the group fairness literature. However, the majority of these works has concentrated predominantly on the development of algorithms leading to models that exhibit similar performance across different clients rather than models that exhibit similar performance across different demographic groups~\cite{Li2020Fair}.

One such approach is agnostic federated learning (AFL) \cite{AFL}, whose aim is to learn a model that optimizes the performance of the worst performing client. 
Another FL approach proposed in \cite{Li2020Fair}, extends AFL by adding an extra fairness constraint to flexibly control performance disparities across clients. Similarly, tilted empirical risk minimization \cite{li2021tilted} uses a hyperparameter called tilt to enable fairness or robustness by magnifying or suppressing the impact of individual client losses, respectively. FedMGDA$+$ is an algorithm that combines minimax optimization coupled with Pareto efficiency \cite{micro_theory} and gradient normalization to ensure fairness across users and robustness against malicious clients. See also other related works in~\cite{ditto}.

{A very recent federated learning work, namely FCFL \cite{DBLP:journals/corr/abs-2108-08435}, focuses on improving the worst performing client while ensuring a certain level of local group fairness by employing gradient-based constrained multi-objective optimization in order to address the proposed challenge. }

Our work concentrates on developing a federated learning algorithm for guaranteeing fairness across all demographic groups included across clients datasets. It therefore naturally departs from existing federated learning approaches such as AFL \cite{AFL}, FedMGDA$+$ \cite{fedmgda+} and $q$-FFL \cite{Li2020Fair} that focus on across-client fairness since, as we prove in Section \ref{clientvsgroupfairness}, a model that guarantees client fairness only guarantees group fairness under some special conditions.

It also departs from FCFL \cite{DBLP:journals/corr/abs-2108-08435}, which considers group fairness to be a per-client objective associated only to the locally available groups. Our primary goal is to learn a model solving (demographic) group fairness across any groups included in the clients distribution, independently of the groups representation in a particular client.

\section{Problem Formulation}\label{sec:prob_form}

\subsection{Group Fairness in Centralized Machine Learning}

We first describe the standard minimax group fairness problem in a centralized machine learning setting \cite{diana2020convergent,martinez2020minimax}, where there is a single entity/server holding all relevant data and responsible for learning a group fair model (see Figure \ref{fig:Fairschemes}). We concentrate on classification tasks, though our approach also applies to other learning tasks such as regression. 
 Let the triplet of random variables $(X,Y,A) \in \mathcal{X} \times \mathcal{Y} \times \mathcal{A}$ represent input features, target, and demographic groups. Let also $p(X,Y,A)=p(A) \cdot p(X,Y|A)$ represent the joint distribution of these random variables where $p(A)$ represents the prior distribution of the different demographic groups and $p(X,Y|A)$ their data conditional distribution.

Let $\ell:\Delta^{|\mathcal{Y}|-1} \times \Delta^{|\mathcal{Y}|-1} \to {\rm I\!R}_+$ be a loss function where $\Delta$ represents the probability simplex.
We now consider that the entity will learn an hypothesis $h$ drawn from an hypothesis class $\mathcal{H}=\{h:\mathcal{X} \rightarrow \Delta^{|\mathcal{Y}|-1} \}$, that solves the optimization problem given by
\begin{equation}\label{objective_centralized}
    \min_{h \in \mathcal{H}} \max_{a \in \mathcal{A}} r_{a}(h)\textnormal{,  } r_a(h)= \mathop{\E}_{(X,Y)\sim p(X,Y|A=a)}[\ell(h(X),Y)|A=a].
\end{equation}
Note that this problem involves the minimization of the expected risk of the worst performing demographic group.

Importantly, under the assumption that the loss is a convex function w.r.t the hypothesis\footnote{This is true for the most common functions in machine learning settings such as Brier score and cross entropy.} and the hypothesis class is a convex set, solving the minimax objective in Eq. \ref{objective_centralized} is equivalent to solving

\begin{equation}\label{objective_centralized_linear}
    \min_{h \in \mathcal{H}} \max_{a \in \mathcal{A}} r_{a}(h) \ge  \min_{h \in \mathcal{H}} \max_{\mu \in \Delta^{|\mathcal{A}|-1}_{\ge \epsilon}} \sum_{a\in\mathcal{A}}\mu_a r_a(h) 
\end{equation}

where $\Delta_{\ge \epsilon}^{|\mathcal{A}| - 1}$ represent the vectors in the simplex with all of their components larger than $\epsilon$. 
Note that if $\epsilon=0$ the inequality in Eq. \ref{objective_centralized_linear} becomes an equality, however, allowing zero value coefficients may lead to models that are weakly, but not strictly, Pareto optimal \cite{geoffrion1968proper,miettinen2012nonlinear}. 

The minimax objective over the linear combination can be achieved by alternating between projected gradient ascent or multiplicative weight updates to optimize the weights, and stochastic gradient descent to optimize the model \cite{chen2018,diana2020convergent,martinez2020minimax}.

\subsection{Group Fairness in Federated Learning}

We now describe our proposed group fairness federated learning problem; this problem differs from the previous one because the data is now distributed across multiple clients but each client (or the server) do not have direct access to the data held by other clients. See also Figure \ref{fig:Fairschemes}.

In this setting, we incorporate categorical variable $K \in \mathcal{K}$ to our data tuple $(X,Y,A,K)$ to indicate the clients participating in the federation. The joint distribution of these variables is $p(X,Y,A,K)=p(K) \cdot p(A|K) \cdot p(X,Y|A,K)$, where $p(K)$ represents a prior distribution over clients -- which in practice is the fraction of samples that are acquired by client $K$ relative to the total number of data samples --, $p(A|K)$, and $p(X,Y|A,K)$ represent the distribution of the groups and the distribution of the input and target variables conditioned on a client. 
We assume that the group-conditional distribution is the same across clients, meaning $p(X,Y|A,K) = p(X,Y|A)$. Note that our model explicitly allows for the distribution of the demographic groups to depend on the client, accommodating for the fact that certain clients may have a higher (or lower) representation of certain demographic groups over others.

We now aim to learn a model $h \in \mathcal{H} $ that solves the minimax fairness problem as presented in Eq. \ref{objective_centralized}, but considering that the group loss estimates are split into $|\mathcal{K}|$ estimators associated to each client. We therefore re-express the linear weighted formulation of Eq. \ref{objective_centralized_linear} using importance weights, allowing to incorporate the role of the different clients, as follows:
\begin{equation}\label{fed_objective_convex}
\begin{array}{ll}
     \min\limits_{h \in \mathcal{H}} \max\limits_{\mu \in \Delta^{|\mathcal{A}|-1}_{\ge \epsilon}} \sum\limits_{a\in\mathcal{A}}\mu_ar_a(h) =& 
    
    \min\limits_{h \in \mathcal{H}} \max\limits_{\mu \in \Delta^{|\mathcal{A}|-1}_{\ge \epsilon}}\sum\limits_{a \in \mathcal{A}} p(A=a) w_a r_a(h) 
   =  \\
     & \min\limits_{h \in \mathcal{H}} \max\limits_{\mu \in \Delta^{|\mathcal{A}|-1}_{\ge \epsilon}}\sum\limits_{k \in \mathcal{K}}  p(K=k) \sum\limits_{a \in \mathcal{A}} p(A=a|K=k) w_a r_a(h)
        =  \\
     & \min\limits_{h \in \mathcal{H}} \max\limits_{\mu \in \Delta^{|\mathcal{A}|-1}_{\ge \epsilon}}\sum\limits_{k \in \mathcal{K}}  p(K=k) r_k(h,{\bm w})
\end{array}
\end{equation}

where $r_k(h,{\bm w})=\sum\limits_{a \in \mathcal{A}} p(A=a|K=k) w_a r_a(h)$ is the expected client risk and $w_a=\mu_a/p(A=a)$ denotes the importance weight for a particular group.

However, there is an immediate non-trivial challenge that arises within this proposed federated learning setting in relation to the centralized one described earlier: we need to devise an algorithm that solves the objective in Eq. \ref{fed_objective_convex} under the constraint that the different clients cannot share their local data with the server or with one another, but -- in line with conventional federated learning settings \cite{DRFA,Li2020Fair,DBLP:journals/corr/McMahanMRA16,AFL}-- only local client updates of a global model (or other quantities such as local risks) are shared with the server.

\section{Client Fairness vs. Group Fairness in Federated Learning}\label{clientvsgroupfairness}

Prior proposing a federated learning algorithm to solve our proposed group fairness problem, we first reflect whether a model that solves the more widely used client fairness objective in federated learning settings given by \cite{AFL}:
\begin{equation}
    \min\limits_{h \in \mathcal{H}}\max\limits_{k \in \mathcal{K}}r_{k}(h) = \min\limits_{h \in \mathcal{H}}\max\limits_{\bm\lambda \in \Delta^{|\mathcal{K}|-1}} \mathop{\E}_{\mathcal{D}_{\bm\lambda}} [\ell(h(X),Y)] \label{client_fairness_objective} 
\end{equation}
where we let $\mathcal{D}_{\bm\lambda}=\sum_{k=1}^{|\mathcal{K}|} \lambda_k p(X,Y|K=k)$ denote a joint data distribution over the clients, also solves our proposed minimax group fairness objective given by:
\begin{equation}
     \min\limits_{h \in \mathcal{H}}\max\limits_{a \in \mathcal{A}}r_{a}(h) = \min\limits_{h \in \mathcal{H}}\max\limits_{\bm\mu \in \Delta^{|\mathcal{A}|-1}} \mathop{\E}_{\mathcal{D}_{\bm\mu}}[\ell(h(X),Y)] \label{group_fairness_objective} 
\end{equation}
where we let $\mathcal{D}_{\bm\mu}=\sum_{a=1}^{|\mathcal{A}|} \mu_a p(X,Y|A=a)$ denote a joint data distribution over sensitive groups.

The following lemma illustrates that a model that is minimax fair with respect to the clients is equivalent to a relaxed minimax fair model with respect to the (demographic) groups.

\begin{sublemma}\label{lemma1}

Let $\rmP_{\mathcal{A}}$ denote a matrix whose entry in row $a$ and column $k$ is $p(A=a|K=k)$ (i.e. the prior of group $a$ in client $k$). 
Then, given a solution to the minimax problem across clients
\begin{equation}
    h^*,\bm\lambda^* \in \arg\min_{h \in \mathcal{H}}\max_{\bm\lambda \in \Delta^{|\mathcal{K}|-1}} \mathop{\E}_{\mathcal{D}_{\bm \lambda}}[\ell(h(X),Y)], 
\end{equation}
$\exists$ $\bm\mu^*=\rmP_{\mathcal{A}}\bm\lambda^*$ that is solution to the following constrained minimax problem across sensitive groups
\begin{equation}
     h^*,\bm\mu^* \in \arg\min_{h \in \mathcal{H}}\max_{\bm \mu \in \rmP_{\mathcal{A}}\Delta^{|\mathcal{K}|-1}}\mathop{\E}_{\mathcal{D}_{\bm \mu}}[\ell(h(X),Y)],
\end{equation}

where the weighting vector $\bm\mu$ is constrained to belong to the simplex subset defined by $\rmP_{\mathcal{A}}\Delta^{|\mathcal{K}|-1} \subseteq \Delta^{|\mathcal{A}|-1}$.
In particular, if the set ${\Gamma} = \big\{ \bm\mu' \in \rmP_{\mathcal{A}}\Delta^{|\mathcal{K}|-1}$: ${\bm\mu'} \in \arg\min\limits_{h \in \mathcal{H}}\max\limits_{ {\bm \mu} \in \Delta^{|\mathcal{A}|-1}}\mathop{\E}_{\mathcal{D}_{\bm \mu}}[\ell(h(X),Y)]\big\} \not= \emptyset$, then ${\bm \mu}^* \in \Gamma $, and the minimax fairness solution across clients is also a minimax fairness solution across demographic groups. 
\end{sublemma}

Lemma \ref{lemma1} proves that being minimax with respect to the clients is equivalent to finding the group minimax model constraining the weighting vectors $\bm\mu$ to be inside the simplex subset $\rmP_{\mathcal{A}}\Delta^{|\mathcal{K}|-1}$. Therefore,
if this set already contains a group minimax weighting vector, then the group minimax model is equivalent to client minimax model.
Another way to interpret this result is that being minimax with respect to the clients is the same as being minimax for any group assignment $\mathcal{A}$ such that linear combinations of the groups distributions are able to generate all clients distributions, and there is a group minimax weighting vector in $\rmP_{\mathcal{A}}\Delta^{|\mathcal{N}|-1}$. 

Being minimax at the client and group level relies on $\rmP_{\mathcal{A}}\Delta^{|\mathcal{K}|-1}$ containing the minimax weighting vector. In particular, if for each sensitive group there is a client comprised entirely of this group ($\rmP_{\mathcal{A}}$ contains a identity block), then $\rmP_{\mathcal{A}}\Delta^{|\mathcal{K}|-1} = \Delta^{|\mathcal{A}|-1}$ and group and client level fairness are guaranteed to be fully compatible. Another trivial example is when at least one of the client's group priors is equal to a group minimax weighting vector. This result also suggests that client level fairness may also differ from group level fairness. This motivates us to develop a new federated learning algorithm to guarantee group fairness that -- where the conditions of the lemma hold -- also results in client fairness. We experimentally validate the insights deriving from Lemma \ref{lemma1} in Section \ref{sec:experiemtns}.

\section{MinMax Group Fairness Federating Learning Algorithm}\label{sec:algo}

We now propose an algorithm -- Federated Minimax (FedMinMax) -- to solve the group fairness problem in Eq. \ref{fed_objective_convex}. 

Clearly, the clients are not able to calculate the statistical averages appearing in Eq. \ref{fed_objective_convex} because the underlying data distributions are unknown. Therefore, we let each client $k$ have access to a dataset $\mathcal{S}_k = \{(x_i^{k},y_i^{k},a_i^{k}); i=1,\ldots,n_k\}$ containing various data points drawn i.i.d according to $p(X,Y,A|K=k)$. We also define three additional sets: (a) $\mathcal{S}_{a,k}=\{(x_i^{k},y_i^{k},a_i^{k}) \in \mathcal{S}_k: a_i=a\}$ is a set containing all data examples associated with group $a$ in client $k$; (b) $\mathcal{S}_{a}=\bigcup\limits_{k \in \mathcal{K}} \mathcal{S}_{k,a}$ is the set containing all data examples associated with group $a$ across the various clients; and (c) $\mathcal{S} = \bigcup\limits_{a \in \mathcal{A}} \mathcal{S}_{a} = \bigcup\limits_{k \in \mathcal{K}} \bigcup\limits_{a \in \mathcal{A}} \mathcal{S}_{a,k}$ is containing all data examples across groups and across clients. 

Note again that -- in view of our modelling assumptions -- it is possible that $\mathcal{S}_{a,k}$ can be empty for some $k$ and some $a$ implying that such a client does not have data realizations for such group.

We will also let the model $h$ be parameterized via a vector of parameters $ \bm \theta \in \Theta$, i.e. $h(\cdot) = h(\cdot;\bm \theta)$. \footnote{This vector of parameters could correspond to the set of weights / biases in a neural network.} Then, one can approximate the relevant statistical risks using empirical risks as follows:

\begin{equation}
\label{empirical_group_risk}
\begin{array}{cc}
     \hat{r}_k(\bm \theta, { \bm w}) = \sum_{a \in \mathcal{A}}\frac{n_{a,k}}{n_k} {\hat w}_a \hat{r}_{a,k}(\bm \theta),& \hat{r}_a (\bm \theta) = \sum_{k \in \mathcal{K}} \frac{n_{a,k}}{n_a} \hat{r}_{a,k}(\bm \theta)  
\end{array}
\end{equation}

where $\hat{r}_{a,k} (\bm \theta) = \frac{1}{n_{a,k}}\sum_{(x,y) \in \mathcal{S}_{a,k}} \ell (h(x;\bm \theta), y)$,  ${\hat w}_a=\mu_a /(n_a/n)$, $n_k = |\mathcal{S}_k|$, $n_a = |\mathcal{S}_a|$, $n_{a,k} = |\mathcal{S}_{a,k}|$, and $n = |\mathcal{S}|$. Note that $\hat{r}_k(\bm \theta, \bm w)$ is an estimate of $r_k(\bm \theta, \bm w)$, $\hat{r}_a(\bm \theta)$ is an estimate of $r_a(\theta)$, and $\hat{r}_{a,k} (\bm \theta)$ is an estimate of $r_{a,k}(\theta)$.

We consider the importance weighted empirical risk ${\hat r}_k$ since the clients do not have access to the data distribution but instead to a dataset with finite samples. 

Therefore, the clients in coordination with the central server attempt to solve the optimization problem given by:
\begin{equation}\label{empirical_fed_objective}
\min_{\bm \theta \in \Theta} \max_{\bm\mu \in \Delta_{\ge \epsilon}^{|
\mathcal{A}| - 1}} \hat{r}_a(\bm \theta)  \coloneqq \sum_{a \in \mathcal{A}} \mu_a \hat{r}_a(\bm \theta)
\textnormal{ or equivalently, }
\min_{\bm \theta \in \Theta} \max_{\bm\mu \in \Delta_{\ge \epsilon}^{|
\mathcal{A}| - 1}}  \sum_{k \in \mathcal{K}} \frac{n_k}{n} \hat{r}_k(\bm \theta, { \bm w}).
\end{equation}

\begin{wrapfigure}{L}{0.53\textwidth}

{\centering
\begin{minipage}{\linewidth}
\begin{algorithm}[H]
\footnotesize
    \caption{\footnotesize\textsc{Federated MiniMax (FedMinMax) }}
    \label{alg:fedminmax}
    {\bfseries Input:} $\mathcal{K}$: Set of clients, $T:$ total number of communication rounds,
     $\eta_{\bm \theta}$: model learning rate, $\eta_{\bm\mu}$: global adversary learning rate, $\mathcal{S}_{a,k}$: set of examples for group $a$ in client $k$, $\forall a \in \mathcal{A}$ and $\forall k \in \mathcal{K}$.
     
    \begin{algorithmic}[1]
    \setstretch{1.35}
    
    \STATE Server {\bfseries initializes} $\bm\mu^0\leftarrow \rho = \{|\mathcal{S}_a|/|\mathcal{S}| \}_{a \in \mathcal{A}}$ and $\bm \theta^0$ randomly.
    
    \FOR{$t=1$ {\bfseries to} $T$}

    \STATE Server {\bfseries computes} $\bm w^{t-1} \leftarrow \bm \mu^{t-1} / \rho$
    
    \STATE Server {\bfseries broadcasts} $\bm \theta^{t-1}$, $\bm w^{t-1}$ 

      \FOR{each client $k \in \mathcal{K}$ {\bfseries in parallel }}
        \STATE  $\bm \theta^{t}_k \leftarrow \bm \theta^{t-1} - \eta_\theta \nabla_\theta \hat{r}_k(\bm \theta^{t-1}, \bm w^{t-1})$%
     
        \STATE Client-$k$ {\bfseries obtains} and {\bfseries sends} $\{\hat{r}_{a,k}(\bm \theta^{t-1})\}_{a \in \mathcal{A}}$ and  $\bm \theta^{t}_k$ to server
        \ENDFOR 
        
    \STATE Server {\bfseries computes}:  $\bm \theta^{t} \leftarrow \sum_{k \in \mathcal{K}} \frac{n_k}{n} \bm \theta^{t}_k$
     
     \STATE Server {\bfseries updates}
     \newline
     $\bm\mu^{t} \leftarrow \prod_{\Delta^{|\mathcal{A}|-1}} \Big (\bm{{\mu}}^{t-1} +\eta_{\bm{\mu}} {\nabla}_{\bm{\mu}}\langle\ \bm \mu^{t-1},\hat{r}_a(\bm \theta^{t-1})\rangle \Big ) $
     \ENDFOR
    
    \end{algorithmic}
    {\bfseries Outputs}: $\frac{1}{T} \sum_{t =1}^T \bm \theta^t$
    \end{algorithm}    
    
    \end{minipage}
\par
}
\end{wrapfigure}

The objective in Eq. \ref{empirical_fed_objective} can be interpreted as a zero-sum game between two players: the learner aims to minimize the objective by optimizing the model parameters $\bm \theta$ and the adversary seeks to maximize the objective by optimizing the weighting coefficients $\bm \mu$.

We use a non-stochastic variant of the stochastic-AFL algorithm introduced in \cite{AFL}. 
Our version, provided in Algorithm \ref{alg:fedminmax}, assumes that all clients are available to participate in each communication round $t$. In particular, in each round $t$, the clients receive the latest model parameters $\bm \theta^{t-1}$, the clients then perform one gradient descent step using all their available data, and the clients then share the updated model parameters along with certain empirical risks with the server. The server (learner) then performs a weighted average of the client model parameters $\bm \theta^{t} = \sum_{k \in \mathcal{K}} \frac{n_k}{n} \bm \theta^{t}_k$. 

The server also updates the weighting coefficient using a projected gradient ascent step in order to guarantee that the weighting coefficient updates are consistent with the constraints. We use the Euclidean algorithm proposed in \cite{10.1145/1390156.1390191} in order to implement the projection operation ($\prod_{\Delta^{|\mathcal{A}|-1}} (\cdot)$).

We can also show that our algorithm can exhibit convergence guarantees.

\begin{sublemma}\label{centralized_vs_federated_solution}
Consider our federated learning setting (Figure \ref{fig:Fairschemes}, right) where each entity $k$ has access to a local dataset $\mathcal{S}_k = \bigcup_{a \in \mathcal{A}} \mathcal{S}_{a,k}$ and a centralized machine learning setting (Figure \ref{fig:Fairschemes}, left) where there is a single entity that has access to a single dataset $\mathcal{S} = \bigcup_{k \in \mathcal{K}} \mathcal{S}_k = \bigcup_{k \in \mathcal{K}} \bigcup_{a \in \mathcal{A}} \mathcal{S}_{a,k}$ (i.e. this single entity in the centralized setting has access to the data of the various clients in the distributed setting).

Then, Algorithm \ref{alg:fedminmax} and Algorithm \ref{alg:centralized_minmax} (in supplementary material, Appendix \ref{more_results}) lead to the same global model provided that learning rates and model initialization are identical.

\end{sublemma}

This lemma shows that our federated learning algorithm inherits any convergence guarantees of existing centralized machine learning algorithms. In particular, assuming that one can model the single gradient descent step using a $\delta$-approximate Bayesian Oracle~\cite{chen2018}, we can show that a centralized algorithm converges and hence our FedMinMax one also converges too (under mild conditions on the loss function, hypothesis class, and learning rates). See Theorem 7 in ~\cite{chen2018}.

\section{Experimental Results}\label{sec:experiemtns}

\begin{wraptable}{l}{0.55\textwidth}

\resizebox{0.55\textwidth}{!}{

\begin{tabular}{llll}
\toprule
 Setting&  Method &     Worst Group Risk  & Best Group Risk\\
\midrule
 ESG & AFL &    0.485$\pm$0.0 &  0.216$\pm$0.001  \\
&FedAvg &    0.487$\pm$0.0 &  0.214$\pm$0.002  \\
& $q$-FedAvg ($q$=0.2) &  0.479$\pm$0.002 &   0.22$\pm$0.002  \\
& $q$-FedAvg ($q$=5.0) &  0.478$\pm$0.002 &  0.223$\pm$0.004  \\
& FedMinMax (ours) &    \textbf{0.451$\pm$0.0} &   \textbf{0.31$\pm$0.001} \\

\midrule
 SSG & AFL &    \textbf{0.451$\pm$0.0} & \textbf{0.31$\pm$0.001} \\
& FedAvg &  0.483$\pm$0.002 &  0.219$\pm$0.001  \\
& $q$-FedAvg ($q$=0.2) &   0.476$\pm$0.001 &  0.221$\pm$0.002  \\
& $q$-FedAvg ($q$=5.0) &  0.468$\pm$0.005 &  0.274$\pm$0.004  \\
& FedMinMax (ours) &    \textbf{0.451$\pm$0.0} &  \textbf{0.309$\pm$0.003} \\

\midrule
\multicolumn{2}{c}{Centalized Minmax Baseline} &
 \textbf{0.451$\pm$0.0} &  \textbf{0.308$\pm$0.001} \\
\bottomrule
\end{tabular}
}\caption{Testing brier score risks for FedAvg, AFL, $q$-FedAvg and FedMinmax across different federated learning scenarios on the synthetic dataset for binary classification involving two sensitive groups. PSG scenario is not included because for $|\mathcal{A}|=2$ it is equivalent to SSG.}

\label{tab:synthetic}
\end{wraptable}

To validate the benefits of the proposed FedMinMax approach, we consider three federated learning scenarios:
(1) \textit{Equal access to Sensitive Groups (ESG)}, where every client has access to all sensitive groups but does not have enough data to train a model individually, to examine a case where group and client fairness are not equivalent; (2) \textit{Partial access to Sensitive Groups (PSG)} where each client has access to a subset of the available groups memberships, to compare the performances when there is low or no local representation of particular groups; (3) \textit{access to a Single Sensitive Group (SSG)}, each client holds data from one sensitive group, for showcasing the group and client fairness objectives equivalence derived from Lemma \ref{lemma1}.

In all experiments we consider a federation consisting of 40 clients and a single server that orchestrates the training procedure per Algorithm 1. We benchmark our approach against AFL \cite{AFL}, $q$-FedAvg \cite{Li2020Fair}, and FedAvg \cite{DBLP:journals/corr/McMahanMRA16}. Further, as a baseline, we also run FedMinMax with one client (akin to centralized ML) to confirm Lemma \ref{centralized_vs_federated_solution}.

We generated a synthetic dataset for binary classification involving two sensitive groups (i.e. $|\mathcal{A}|=2$), details available in Appendix \ref{more_results}. We provide the performance on ESG and SSG scenarios\footnote{Note that PSG scenario is valid only for datasets where $|\mathcal{A}|>2$, else its equivalent to SSG setting.} in Table \ref{tab:synthetic}. FedMinMax performs similarly to Centralized Minmax Baseline for both sensitive groups in all setups, as proved in Lemma \ref{centralized_vs_federated_solution}. AFL yields the same solution as FedMinMax and Centralized Minmax Baseline only in SSG where group fairness is implied by client fairness. 
Both FedAvg and $q$-FedAvg fail to achieve minimax group fairness.

\begin{wraptable}{r}{0.6\textwidth}
\resizebox{0.6\textwidth}{!}{
\begin{tabular}{lllll}
\toprule
 Setting&  Method &T-shirt  &  Pullover &  Shirt \\
\midrule
 ESG &AFL & 0.239$\pm$0.003 &0.262$\pm$0.001 &  0.494$\pm$0.004 \\
&FedAvg &  0.243$\pm$0.003& 0.262$\pm$0.001 &  0.492$\pm$0.003\\
&FedMinMax (ours) &  0.261$\pm$0.006 &  0.256$\pm$0.027 & \textbf{0.307$\pm$0.01}  \\
\midrule
SSG &AFL &  0.267$\pm$0.009 &  0.236$\pm$0.013 &   \textbf{0.307$\pm$0.003} \\
&FedAvg &  0.227$\pm$0.003 & 0.236$\pm$0.004 &   0.463$\pm$0.003 \\
&FedMinMax (ours) &  0.269$\pm$0.012 &  0.238$\pm$0.017 & \textbf{0.309$\pm$0.011}\\
\midrule
PSG & AFL & 0.244$\pm$0.007 &   0.257$\pm$0.066  &  0.425$\pm$0.019  \\
&FedAvg &  0.229$\pm$0.008  & 0.236$\pm$0.004 & 0.464$\pm$0.011   \\
&FedMinMax (ours) &  0.263$\pm$0.013 &  0.228$\pm$0.011 & \textbf{0.31$\pm$0.008 } \\
\midrule
\multicolumn{2}{c}{Centalized Minmax Baseline} &  0.259$\pm$0.01 & 0.239$\pm$0.051 &  \textbf{0.311$\pm$0.006} \\
\bottomrule
\end{tabular}

}\caption{Testing brier score risks for the top-3 worst groups in FashionMNIST dataset. The risks for all the available classes are available in Table \ref{tanalytic_fmnist}, in Appendix \ref{more_results}. }

\label{tab:fmnist}
\end{wraptable}

For FashionMNIST we use all ten clothing target categories, which we assign both as targets and sensitive groups (i.e. $|\mathcal{A}|=10$). In Table \ref{tab:fmnist} we examine the performance on the three worst categories: \textit{T-shirt}, \textit{Pullover}, and \textit{Shirt} (the risks for all classes are available in Table \ref{tanalytic_fmnist}, in Appendix \ref{more_results}). In all settings, FedMinMax improves the risk of the worst group, \textit{Shirt}, more than it degrades the performance on the \textit{T-shirt} class, all while maintaining the same risk on \textit{Pullover} as FedAvg. AFL performs akin to FedMinMax for the SSG setup but not on the other settings as expected by Lemma \ref{lemma1}. Note that FedMinMax has the best worst group performance in all settings as expected.
More details about datasets, models, experiments and results are provided in Appendix \ref{more_results}.

\section{Conclusion}
In this work, we formulate (demographic) group fairness in federated learning setups where different participating entities may only have access to a subset of the population groups during the training phase (but not necessarily the testing phase), exhibiting minmax fairness performance guarantees akin to those in centralized machine learning settings.

We formally show how our fairness definition differs from the existing fair federated learning works, offering conditions under which conventional client-level fairness is equivalent to group-level fairness.
We also provide an optimization algorithm, FedMinMax, to solve the minmax group fairness problem in federated setups that exhibits minmax guarantees akin to those of minmax group fair centralized machine learning algorithms.

We empirically confirm that our method outperforms existing federated learning methods in terms of group fairness in various learning settings and validate the conditions that the competing approaches yield the same solution as our objective.

\label{sec:conclusion}

\newpage

\bibliographystyle{plain}
\bibliography{main.bib}

\newpage

\appendix

\section{Appendix: Proofs}\label{proofs_sup}

{
\begin{lemma}\textbf{\textsc{\ref{lemma1}}} 
Let $\rmP_{\mathcal{A}}$ denote a matrix whose entry in row $a$ and column $k$ is $p(A=a|K=k)$ (i.e. the prior of group $a$ in client $k$). 
Then, given a solution to the minimax problem across clients
\begin{equation}
    h^*,\bm\lambda^* \in \arg\min\limits_{h \in \mathcal{H}}\max\limits_{\bm\lambda \in \Delta^{|\mathcal{K}|-1}} \mathop{\E}\limits_{\mathcal{D}_{\bm \lambda}}[\ell(h(X),Y)], 
\end{equation}
$\exists$ $\bm\mu^*=\rmP_{\mathcal{A}}\bm\lambda^*$ that is solution to the following constrained minimax problem across sensitive groups
\begin{equation}
     h^*,\bm\mu^* \in \arg\min\limits_{h \in \mathcal{H}}\max\limits_{\bm \mu \in \rmP_{\mathcal{A}}\Delta^{|\mathcal{K}|-1}}\mathop{\E}\limits_{\mathcal{D}_{\bm \mu}}[\ell(h(X),Y)],
\end{equation}
where the weighting vector $\bm\mu$ is constrained to belong to the simplex subset defined by $\rmP_{\mathcal{A}}\Delta^{|\mathcal{K}|-1} \subseteq \Delta^{|\mathcal{A}|-1}$.
In particular, if the set ${\Gamma} = \big\{ \bm\mu' \in \rmP_{\mathcal{A}}\Delta^{|\mathcal{K}|-1}$: ${\bm\mu'} \in \arg\min\limits_{h \in \mathcal{H}}\max\limits_{ {\bm \mu} \in \Delta^{|\mathcal{A}|-1}}\mathop{\E}\limits_{\mathcal{D}_{\bm \mu}}[\ell(h(X),Y)]\big\} \not= \emptyset$, then ${\bm \mu}^* \in \Gamma $, and the minimax fairness solution across clients is also a minimax fairness solution across demographic groups. 
\end{lemma}


\begin{proof} The objective for optimizing the global model for the worst mixture of client distributions is:
\begin{equation}
\label{afl_objective}
\min\limits_{h \in \mathcal{H}} \max\limits_{\boldsymbol\lambda \in \Delta^{|\mathcal{K}|-1} }
\mathop{\E}\limits_{\mathcal{D}_{\boldsymbol\lambda}}[l(h(X),Y)] = \min\limits_{h \in \mathcal{H}} \max\limits_{\boldsymbol\lambda \in \Delta^{|\mathcal{K}|-1}} \sum\limits_{k=1}^{|\mathcal{K}|} \lambda_k\mathop{\E}\limits_{\mathcal{D}_k}[l(h(X),Y)],
\end{equation}

given that $\mathcal{D}_{\bm\lambda}=\sum\limits_{k=1}^{|\mathcal{K}|} \lambda_k p(X,Y|K=k)$. 
 Since $p(X,Y|K=k)=\sum\limits_{a \in \mathcal{A}}p(A=a|K=k)p(X,Y|A)$ 
 with $p(A=a|K=k)$ being the prior of $a \in \mathcal{A}$ for client $k$, and $p(X,Y|A=a)$ is the distribution conditioned on the sensitive group $a \in \mathcal{A}$, Eq. (\ref{afl_objective}) can be re-written as:
\begin{equation}
\label{clients_groups_equivalence}
\begin{array}{l}

    \min\limits_{h \in \mathcal{H}} \max\limits_{\boldsymbol\lambda \in \Delta^{|\mathcal{K}|-1}} \sum\limits_{k=1}^{|\mathcal{K}|} \lambda_k \sum\limits_{a \in \mathcal{A}} p(A=a|K=k)\underset{p(X,Y|A=a)}{\mathop{\E}}[l(h(X),Y)] = \\
    \\
    \min\limits_{h \in \mathcal{H}} \max\limits_{\boldsymbol\lambda \in \Delta^{|\mathcal{K}|-1}} \sum\limits_{a \in \mathcal{A}}\underset{p(X,Y|A=a)}{\mathop{\E}}[l(h(X),Y)] \Big ( \sum\limits_{k=1}^{|\mathcal{K}|} p(A=a|K=k)\lambda_k \Big ) = \\
    \\
        \min\limits_{h \in \mathcal{H}} \max\limits_{\boldsymbol\mu \in \rmP_\mathcal{A} \Delta^{|\mathcal{K}|-1}} \sum\limits_{a \in \mathcal{A}} \mu_a\underset{p(X,Y|A=a)}{\mathop{\E}}[l(h(X),Y)].
\end{array}
\end{equation}

Where we defined $\mu_a= \sum\limits_{k=1}^{|\mathcal{K}|} p(A=a|K=k)\lambda_k$, $\forall a \in \mathcal{A}$, this creates the vector $\boldsymbol\mu=\rmP_\mathcal{A}\boldsymbol\lambda \subseteq \rmP_\mathcal{A} \Delta^{|\mathcal{K}|-1}$. It holds that the set of possible $\mu$ vectors satisfies $ \rmP_\mathcal{A} \Delta^{|\mathcal{K}|-1} \subseteq \Delta^{|\mathcal{A}|-1}$, since $\rmP_{\mathcal{A}}= \big \{\{p(A=a|K=k)\}_{a \in \mathcal{A}} \big \}_{ k \in \mathcal{K}} \in  {\rm I\!R}_+^{|\mathcal{A}|\times |\mathcal{K}|}$, with $\underset{a \in \mathcal{A}}{\sum}p(A=a|K=k)=1$ $\forall k$ and $\boldsymbol\lambda \in \Delta^{|\mathcal{K}|-1}$. 

Then, from the equivalence in Equation \ref{clients_groups_equivalence} we have that, given any solution
\begin{equation}
\label{eq:lambda_solution}
    h^*,\boldsymbol\lambda^* \in \arg\min\limits_{h \in \mathcal{H}}\max\limits_{\boldsymbol\lambda \in \Delta^{|\mathcal{K}|-1}} \mathop{\E}\limits_{\mathcal{D}_{\bm\lambda}}[\ell(h(X),Y)], 
\end{equation}
then $\boldsymbol\mu^*=\rmP_{\mathcal{A}}\boldsymbol\lambda^*$ is solution to 
\begin{equation}
\label{eq:gamma_solution}
h^*,\boldsymbol\mu^* \in \arg\min\limits_{h \in \mathcal{H}}\max\limits_{\mu \in \rmP_{\mathcal{A}}\Delta^{|\mathcal{K}|-1}}\mathop{\E}\limits_{\mathcal{D}_{\bm\mu}}[\ell(h(X),Y)],
\end{equation}
and
\begin{equation}
\mathop{\E}\limits_{\mathcal{D}_{\bm\mu^*}}[\ell(h^*(X),Y)] = \mathop{\E}\limits_{\mathcal{D}_{\bm\lambda^*}}[\ell(h^*(X),Y)].
\end{equation}

In particular, if the space defined by $\rmP_{\mathcal{A}}\Delta^{|\mathcal{K}|-1}$ contains any  group minimax fair weights, meaning that the set ${\Gamma} = \big\{ \boldsymbol\mu' \in \rmP_{\mathcal{A}}\Delta^{|\mathcal{K}|-1}$: ${\boldsymbol\mu'} \in \arg\min\limits_{h \in \mathcal{H}}\max\limits_{{\bm\mu} \in \Delta^{|\mathcal{A}|-1}}\mathop{\E}\limits_{\mathcal{D}_{\bm\mu}}[\ell(h(X),Y)]\big\}$ is not empty, then it follows that any $\bm \mu^*$ (solution to Equation \ref{eq:gamma_solution}) is already minimax fair with respect to the groups ${\bm\mu}^* \in \Gamma$. And therefore the client-level minimax solution is also a minimax solution across sensitive groups.

\end{proof}
}

{\begin{lemma}\textbf{\textsc{\ref{centralized_vs_federated_solution}}}
Consider our federated learning setting (Figure \ref{fig:Fairschemes}, right) where each entity $k$ has access to a local dataset $\mathcal{S}_k = \bigcup_{a \in \mathcal{A}} \mathcal{S}_{a,k}$ and a centralized machine learning setting (Figure \ref{fig:Fairschemes}, left) where there is a single entity that has access to a single dataset $\mathcal{S} = \bigcup_{k \in \mathcal{K}} \mathcal{S}_k = \bigcup_{k \in \mathcal{K}} \bigcup_{a \in \mathcal{A}} \mathcal{S}_{a,k}$ (i.e. this single entity in the centralized setting has access to the data of the various clients in the distributed setting).

Then, Algorithm \ref{alg:fedminmax} and Algorithm \ref{alg:centralized_minmax} (in supplementary material, Appendix \ref{more_results}) lead to the same global model provided that learning rates and model initialization are identical.
\end{lemma}

\begin{proof} 
 We will show that FedMinMax, in Algorithm \ref{alg:fedminmax} is equivalent to the centralized algorithm, in Algorithm \ref{alg:centralized_minmax} under the following conditions:
 \begin{enumerate}

     \item the dataset on client $k$, in FedMinMax is $\mathcal{S}_k = \bigcup_{a \in \mathcal{A}} \mathcal{S}_{a,k}$ and the dataset in centralized MinMax is $\mathcal{S} = \bigcup_{k \in \mathcal{K}} \mathcal{S}_k = \bigcup_{k \in \mathcal{K}} \bigcup_{a \in \mathcal{A}} \mathcal{S}_{a,k}$, and
     
    \item the model initialization $\theta^0$, the number of adversarial rounds $T$,\footnote{In the federated Algorithm \ref{alg:fedminmax}, we also refer to the adversarial rounds as communication rounds.}, learning rate for the adversary $\eta_\mu$ and learning rate for the learner $\eta_\theta$, are identical for both algorithms.
 \end{enumerate}

This can then be immediately done by showing that steps lines 3-7 in Algorithm 1 are entirely equivalent to step 3 in Algorithm \ref{alg:centralized_minmax}. In particular, note that we can write:
\begin{equation*}
\begin{array}{ll}
\hat{r}(\theta, \bm \mu) & =\sum\limits_{a \in \mathcal{A}}\mu_a\hat{r}_a(\bm \theta)\\

\\&= \sum\limits_{a \in \mathcal{A}}\mu_a \sum\limits_{k \in \mathcal{K}}\frac{n_{a,k}}{n_a}\hat{r}_{a,k}(\bm \theta)\\

\\&= \sum\limits_{a \in \mathcal{A}}\mu_a\frac{n}{n_a}\frac{1}{n} \sum\limits_{k \in \mathcal{K}}n_{a,k}\hat{r}_{a,k}(\bm \theta)
\\

\\&= \sum\limits_{a \in \mathcal{A}} w_a \frac{1}{n} \sum\limits_{k \in \mathcal{K}} \frac{n_{a,k}}{n_k} n_k \hat{r}_{a,k}(\bm \theta)
\\

\end{array}
\end{equation*}

\begin{equation}
\begin{array}{ll}\label{equivalence}
\\&= \sum\limits_{k \in \mathcal{K}} \frac{n_k}{n} \sum\limits_{a \in \mathcal{A}} w_a \frac{n_{a,k}}{n_k} \hat{r}_{a,k}(\bm \theta)
\\
\\&= \sum\limits_{k \in \mathcal{K}} \frac{n_k}{n} \hat{r}_{k}(\bm \theta, \bm w)
\\
\end{array}
\end{equation}

because

\begin{equation}\label{grouprisk}
\hat{r}_k(\bm \theta, \bm w) = \sum\limits_{a \in \mathcal{A}}\frac{n_{a,k}}{n_k} w_a \hat{r}_{a,k}(\bm \theta)\textnormal{, with } w_a= \frac{\mu_a}{\frac{n_a}{n}}, \textnormal{ and }
\hat{r}_a (\bm \theta) = \sum\limits_{k \in \mathcal{K}} \frac{n_{a,k}}{n_a} \hat{r}_{a,k}(\bm \theta).
\end{equation}

Therefore, the following model update:
\begin{equation}
    \bm \theta^{t}= \sum\limits_{k \in \mathcal{K}} \frac{n_k}{n} \bm \theta^{t}_k  =  \sum\limits_{k \in \mathcal{K}} \frac{n_k}{n}  \big(\bm \theta^{t-1} - \eta_\theta \nabla_\theta \hat{r}_k(\bm \theta^{t-1}, \bm w^{t-1}) \big)
\end{equation}
associated with step in 7, at round $t$ of Algorithm 1, is entirely equivalent to the model update
\begin{equation}
    \bm \theta^{t} = \bm \theta^{t-1} - \eta_\theta \nabla_\theta \hat{r}(\bm \theta^{t-1}, \bm w^{t-1})
\end{equation}
associated with step in line 3 at round $t$ of Algorithm \ref{alg:centralized_minmax}, provided that ${\bm \theta}^{t-1}$ is the same for both algorithms. 

It follows therefore by induction that, provided the initialization ${\bm \theta}^{0}$ and learning rate $\eta_\theta$ are identical in both cases the algorithms lead to the same model.
Also, from Eq. \ref{grouprisk}, we have that the projected gradient ascent step in line 4 of Algorithm \ref{alg:centralized_minmax} is equivalent to the step in line 10 of Algorithm \ref{alg:fedminmax}.
\end{proof}}

\section{Appendix: Experiments}\label{more_results}

 \subsection{Experimental Details}\label{experimetnal_details}

\begin{figure}[ht]
    \centering
    \includegraphics[width=13cm]{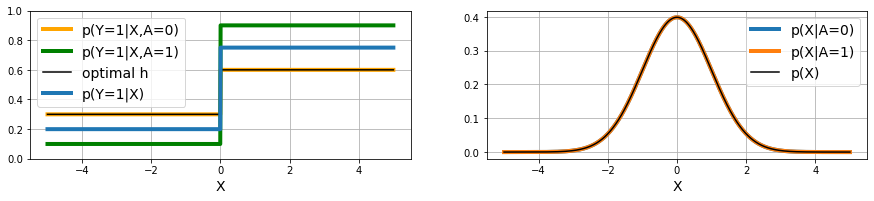}
    \caption{Illustration of the optimal hypothesis $h$ and the conditional distributions $p(Y|X)$ and $p(X|A)$ for the generated synthetic dataset. }
    \label{fig:synthetic_data}
\end{figure}

\paragraph{Datasets.}
For the experiments we use the following datasets:
\begin{itemize}
    \item \textbf{Synthetic.} Let $\mathcal{N}$ and $Ber$ be the  normal and Bernoulli distributions. The data were generated assuming the group variable $A \sim Ber(\frac{1}{2})$, the input features variable $X \sim \mathcal{N}(0,1)$ and the target variable $Y|X,A=a\sim Ber(h^*_a)$, where $h^*_a=u^l_a \mathbbm{1}[x \leq0]+u_a^h \mathbbm{1}[x >0]$ is the optimal hypothesis for group $A=a$. We select $\{u_0^h,u_1^h,u_0^l,u_1^l\}=\{0.6,0.9,0.3,0.1\}$. As illustrated in Figure \ref{fig:synthetic_data}, left side, the optimal hypothesis $h$ is equal to the optimal model for group $A=0$. 
    
    \item \textbf{FashionMNIST.} FashionMNIST is a grayscale image dataset which includes $60,000$ training images and 10,000 testing images. The images consist of $28\times28$ pixels and are classified into 10 clothing categories. In our experiments we consider each of the target categories to be a sensitive group too.

\end{itemize}

\paragraph{Experimental Setting and Model Architectures.} For all the datasets, we use three-fold cross validation to compute the means and standard deviations of the accuracies and risks, with different random initializations. 
Also note that each client's data is unique, meaning that there are no duplicated examples across clients. We assuming that every client is available to participate at each communication round for every method.
 For $q$-FedAvg we use $q=\{0.2,5.0\}$. The learning rate of the classifier for all methods is $\eta_\theta=0.1$ and for the adversary in AFL and FedMinMax we use $\eta_\mu=\eta_\lambda=0.1$.  
The local iterations for FedAvg and $q$-FedAvg are $E=15$. For AFL and FedMinMax the batch size is equal to the number of examples per client while for FedAvg and $q$-FedAvg is equal to $100$.  
For the synthetic dataset, we use an MLP architecture consisting of four hidden layers of size 512 and in the experiments for FashionMNIST we used a CNN architecture with two 2D convolutional layers with kernel size 3, stride 1 and padding 1. Each convolutional layer is followed with a maxpooling layer with kernel size 2, stride 2, dilation 1 and padding 0. All models were trained using Brier score loss function. A summary of the experimental setup is provided in Table \ref{tab:training_settings}.

\begin{table}[ht]
\resizebox{0.99\textwidth}{!}{
\begin{tabular}{llllllll}
\toprule
{Setting }  &  Method &  $\eta_\theta$  &  Batch Size &  Loss & {Hypothesis Type} & Epochs & $\eta_\mu$ or $\eta_\lambda$  \\
\midrule
{ESG,SSG}   & AFL &  0.1& $n_k$& Brier Score & MLP&-&0.1 \\
{ESG,SSG} &FedAvg & 0.1  & 100  & Brier Score & MLP&15& - \\
{ESG,SSG} &$q$-FedAvg & 0.1  & 100  & Brier Score & MLP&15& -  \\
{ESG,SSG} &FedMinMax (ours) & 0.1 & $n_k$  & Brier Score &MLP&- &0.1 \\
{ESG,SSG} &{Centalized Minmax} & 0.1&$n_k$ &Brier Score &MLP &-& 0.1 \\
\midrule
{ESG,SSG,PSG }& AFL &0.1 & $n_k$& Brier Score&  CNN&- & 0.1\\
{ESG,SSG,PSG}&FedAvg &  0.1 & 100 & Brier Score  & CNN&15 & -\\
{ESG,SSG,PSG}&FedMinMax (ours) & 0.1 &  $n_k$ &Brier Score&CNN&-& 0.1\\
{ESG,SSG,PSG}&{Centalized Minmax} & 0.1& $n_k$&Brier Score&CNN&-&0.1 \\
\bottomrule
\end{tabular}
}
\caption{Summary of parameters used in the training process for all experiments. Epochs refers to the local iterations performed at each client,$n_k$ is the number of local data examples in client $k$, $\eta_\theta$ is the model's learning rate and $\eta_\mu$ or $\eta_\lambda$ is the adversary learning rates.}
\label{tab:training_settings}
\end{table}

 \paragraph{Software \& Hardware.} The proposed algorithms and experiments are written in Python, leveraging PyTorch \cite{NEURIPS2019_9015}. The experiments were realised using 1 $\times$ NVIDIA Tesla V100 GPU.

\subsection{Additional Results}
\paragraph{Experiments on FashionMNIST.}
 
 In the \textit{Partial access to Sensitive Groups (PSG)} setting, we distribute the data across 40 participants, 20 of which have access to groups \textit{T-shirt}, \textit{Trouser}, \textit{Pullover}, \textit{Dress} and \textit{Coat} and the other 20 have access to \textit{Sandal}, \textit{Shirt}, \textit{Sneaker}, \textit{Bag} and \textit{Ankle Boot}. The data distribution is unbalanced across clients since the size of local datasets differs among clients (i.e. $n_i\neq n_j \forall i,j \in \mathcal{K},i \neq j$).
 In the \textit{Equal access to Sensitive Groups (ESG)} setting, the 10 classes are equally distributed across the clients, creating a scenario where each client has access to the same amount of data examples and groups (i.e. $n_i= n_j \forall i,j \in \mathcal{K},i \neq j$ and $n_{a,i}= n_{a,j} \forall i,j \in \mathcal{K}, a \in \mathcal{A}, i \neq j$).
 Finally, in the \textit{Single access to Sensitive Groups (SSG)} setting, every client owns only one sensitive group and each group is distributed to only 4 clients. Again, the local datasets are different, $n_i \neq n_j \forall i,j \in \mathcal{K},i \neq j$, creating an unbalanced data distribution.

\begin{figure}[ht]
    \centering
    \includegraphics[width=13cm]{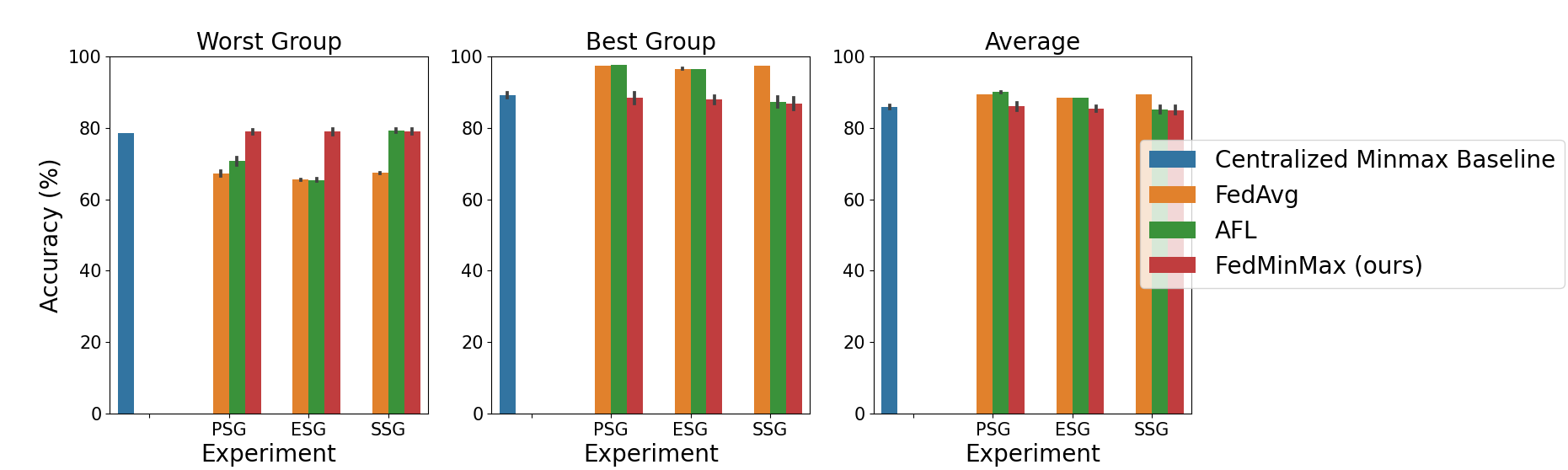}
    \setlength{\abovecaptionskip}{-1pt}
    \caption{Worst Group, Best Group and Average accuracies for AFL, FedAvg and FedMinmax across different federated learning scenarios on the FashionMNIST dataset.}
    \label{fig:fmnist_}
\end{figure}

 We show a comparison of the worst group \textit{Shirt}, the best group \textit{Trousers} and the average accuracies in Figure \ref{fig:fmnist_}.  FedMinMax enjoys a similar accuracy to the Centralized Minmax Baseline, as expected. AFL has similar performance FedMinMax in SSG, where across client fairness implies group fairness, in line with Lemma 1, and FedAvg has similar worst, best and average accuracy, across federated settings. An extended version of group risks is shown in Table \ref{tanalytic_fmnist}.

\begin{table}[ht]
\resizebox{\textwidth}{!}{
\begin{tabular}{llllllllllll}
\toprule
 Setting&  Method &T-shirt  & Trouser & Pullover & Dress & Coat & Sandal & Shirt & Sneaker & Bag & Ankle boot\\
\midrule
 ESG &AFL &  0.239$\pm$0.003 &    0.046$\pm$0.0 &  0.262$\pm$0.001 &  0.159$\pm$0.001 &  0.252$\pm$0.004 &     0.06$\pm$0.0 &  0.494$\pm$0.004 &  0.067$\pm$0.001 &    0.049$\pm$0.0 &   0.07$\pm$0.001 \\
&FedAvg &  0.243$\pm$0.003 &    0.046$\pm$0.0 &  0.262$\pm$0.001 &  0.158$\pm$0.003 &  0.253$\pm$0.002 &    0.061$\pm$0.0 &  0.492$\pm$0.003 &    0.068$\pm$0.0 &    0.049$\pm$0.0 &    0.069$\pm$0.0 \\
&FedMinMax (ours) &  0.261$\pm$0.006 &  0.191$\pm$0.016 &  0.256$\pm$0.027 &  0.217$\pm$0.013 &  0.223$\pm$0.031 &  0.207$\pm$0.027 &   \textbf{0.307$\pm$0.01} &  0.172$\pm$0.016 &  0.193$\pm$0.021 &  0.156$\pm$0.011 \\
\midrule
SSG &AFL &  0.267$\pm$0.009 &  0.194$\pm$0.023 &  0.236$\pm$0.013 &  0.226$\pm$0.012 &  0.262$\pm$0.012 &  0.201$\pm$0.026 &  \textbf{0.307$\pm$0.003} &  0.178$\pm$0.033 &  0.205$\pm$0.025 &  0.162$\pm$0.021 \\
            &FedAvg &  0.227$\pm$0.003 &  0.039$\pm$0.001 &  0.236$\pm$0.004 &  0.143$\pm$0.003 &  0.232$\pm$0.003 &  0.051$\pm$0.001 &  0.463$\pm$0.003 &    0.067$\pm$0.0 &    0.041$\pm$0.0 &  0.063$\pm$0.001 \\
&FedMinMax (ours) &  0.269$\pm$0.012 &    0.2$\pm$0.026 &  0.238$\pm$0.017 &  0.231$\pm$0.013 &  0.252$\pm$0.034 &    0.2$\pm$0.024 &  \textbf{0.309$\pm$0.011} &   0.177$\pm$0.03 &  0.205$\pm$0.032 &  0.169$\pm$0.013 \\
\midrule
PSG & AFL &  0.244$\pm$0.007 &  0.032$\pm$0.001 &  0.257$\pm$0.066 &  0.122$\pm$0.006 &  0.209$\pm$0.098 &  0.045$\pm$0.002 &  0.425$\pm$0.019 &  0.059$\pm$0.001 &  0.041$\pm$0.001 &  0.062$\pm$0.001 \\
&FedAvg &  0.229$\pm$0.008 &    0.039$\pm$0.0 &  0.236$\pm$0.004 &  0.142$\pm$0.002 &  0.232$\pm$0.003 &  0.052$\pm$0.001 &  0.464$\pm$0.011 &  0.067$\pm$0.001 &  0.042$\pm$0.001 &  0.063$\pm$0.001 \\
&FedMinMax (ours) &  0.263$\pm$0.013 &  0.177$\pm$0.026 &  0.228$\pm$0.011 &   0.21$\pm$0.019 &  0.238$\pm$0.025 &   0.182$\pm$0.03 &   \textbf{0.31$\pm$0.008} &   0.16$\pm$0.027 &  0.184$\pm$0.031 &  0.154$\pm$0.018 \\
\midrule
\multicolumn{2}{c}{Centalized Minmax Baseline} &  0.259$\pm$0.01 &  0.173$\pm$0.015 &  0.239$\pm$0.051 &  0.213$\pm$0.008 &  0.24$\pm$0.063 &  0.182$\pm$0.024 & \textbf{0.311$\pm$0.006} &  0.168$\pm$0.018 &  0.18$\pm$0.013 &  0.151$\pm$0.012 \\

\bottomrule
\end{tabular}}
\caption{Brier score risks for FedAvg, AFL and FedMinmax across different federated learning settings on FashionMNIST dataset. Extension of Table \ref{tab:fmnist}.
}
\label{tanalytic_fmnist} 
\end{table}

 \newpage
 \subsection{Centralized MinMax Algorithm}
 We provide the centralized version of FedMinMax in Algorithm \ref{alg:centralized_minmax}.
 
\begin{wrapfigure}{l}{0.99\textwidth}
{\centering
\begin{minipage}{0.7\textwidth}
\begin{algorithm}[H]

    \caption{\textsc{Centralized MinMax Baseline}}
    \label{alg:centralized_minmax}
    {\bfseries Input:} $T:$ total number of adversarial rounds,
     $\eta_{\bm \theta}$: model learning rate, $\eta_{\bm\mu}$: adversary learning rate, $\mathcal{S}_{a}$: set of examples for group $a$, $\forall a \in \mathcal{A}$. 
    \begin{algorithmic}[1]
    \setstretch{1.35}
    
    \STATE Server {\bfseries initializes} $\bm\mu^0\leftarrow \{|\mathcal{S}_a|/|\mathcal{S}| \}_{a \in \mathcal{A}}$ and $\bm \theta^0$ randomly.
    
    \FOR{$t=1$ {\bfseries to} $T$}

    \STATE Server {\bfseries computes}  $\bm \theta^{t}_k \leftarrow \bm \theta^{t-1} - \eta_\theta \nabla_\theta \hat{r}(\bm \theta^{t-1}, \bm \mu^{t-1})$ 
    
     \STATE Server {\bfseries updates}
     \newline
     $\bm\mu^{t} \leftarrow \prod_{\Delta^{|\mathcal{A}|-1}} \Big (\bm{{\mu}}^{t-1} +\eta_{\bm{\mu}} {\nabla}_{\bm{\mu}}\langle\ \bm \mu^{t-1},\hat{r}_a(\bm \theta^{t-1})\rangle \Big ) $
     \ENDFOR
    
    \end{algorithmic}
    {\bfseries Outputs}: $\frac{1}{T} \sum_{t =1}^T \bm \theta^t$
    \end{algorithm}    
    
    \end{minipage}
\par
}
\end{wrapfigure}

\end{document}